\documentclass[sigconf]{acmart}

\usepackage{booktabs}
\usepackage{multirow}
\usepackage{amsmath}

\usepackage{amssymb}

\copyrightyear{2026}
\acmYear{2026}
\setcopyright{cc}
\setcctype{by}
\acmConference[SIGSPATIAL '26]{The 34th ACM International Conference on Advances in Geographic Information Systems}{November 03--06, 2026}{Riverside, CA, USA}
\acmBooktitle{The 34th ACM International Conference on Advances in Geographic Information Systems (SIGSPATIAL '26), November 03--06, 2026, Riverside, CA, USA}
\acmDOI{10.1145/3841645.3842996}
\acmISBN{979-8-4007-2950-8/2026/11}

\title{ScaleTraj: Scale-Decomposed Human Trajectory Generation}

\author{Yulin Zeng}
\affiliation{%
  \institution{Santa Clara University}
  \city{Santa Clara}
  \state{CA}
  \country{USA}}
\email{yzeng@scu.edu}
\author{Yiheng Zhao}
\affiliation{%
  \institution{Concordia University}
  \city{Montreal}
  \country{Canada}}
\email{yiheng.zhao@mail.concordia.ca}
\author{Jun Yan}
\affiliation{%
  \institution{Concordia University}
  \city{Montreal}
  \country{Canada}}
\email{jun.yan@concordia.ca}
\author{Yuhong Liu}
\authornote{Corresponding author.}
\affiliation{%
  \institution{Santa Clara University}
  \city{Santa Clara}
  \state{CA}
  \country{USA}}
\email{yhliu@scu.edu}



\begin{document}

\begin{abstract}
Individual-level trajectories are essential for applications that rely on user-level mobility behavior. Existing mobility-generation approaches face a fundamental tension between individual-level and population-level statistical fidelity: top-down models reproduce aggregate mobility statistics but struggle to guarantee realistic individual trajectories, while bottom-up generators synthesize individual trajectories yet often fail to preserve key population-level mobility regularities after aggregation. In particular, many bottom-up generated populations fail to reproduce the displacement distribution, whose heavy tail is well summarized by a truncated power law. We formalize this phenomenon as the \textbf{Population-Level Displacement Gap}. To investigate its prevalence, we conduct a systematic evaluation of representative model-driven and data-driven trajectory generators. Our results provide evidence that the Population-Level Displacement Gap is prevalent across diverse generation paradigms. To diagnose this gap, we introduce the \textbf{Scale Fidelity Profile (SFP)}, a scale-decomposed metric that quantifies discrepancies in trip allocations across spatial scales. SFP indicates that population-level displacement mismatches co-occur with systematic errors in how trips are allocated across spatial scales. Motivated by this observation, we develop \textbf{ScaleTraj}, an autoregressive Transformer that decomposes location prediction into a Scale Selection Head that decides the movement scale and a Scale-Aware Cell Selection Head that predicts the destination conditioned on the selected scale. Experiments on two datasets show that ScaleTraj reduces the Population-Level Displacement Gap versus model-driven and data-driven baselines, while preserving fidelity. Our code is available at https://github.com/SCU-TrustworthyAI/ScaleTraj.
\end{abstract}

\maketitle

\section{Introduction}
\label{sec:intro}

Granular, individual-level trajectory generation is essential for critical downstream applications, including electric-vehicle charging load prediction and epidemic simulation, which heavily rely on user-level behavioral analytics. While macro-level, top-down generative approaches excel at capturing aggregate population statistics (i.e., often reproducing the characteristic power-law tail with high accuracy), they inherently lack the capacity to satisfy these fine-grained individual analytic tasks. Consequently, researchers rely on bottom-up trajectory generators to synthesize individual traces, bypassing the privacy constraints and prohibitive collection costs.

However, bottom-up models often struggle to sustain population-level statistical properties upon aggregation. Specifically, the aggregated synthetic population can deviate significantly from the corresponding real-data displacement distribution. This paper formalizes this distributional mismatch as the \textbf{Population-Level Displacement Gap}, defined as the statistical discrepancy between the \emph{dataset's own} observed displacement distribution and the distribution obtained after aggregating generated individual trajectories. Literature-consensus values are reported only as external context. Such deviations are not merely statistical artifacts. The displacement distribution governs the frequency of long-range movements, which directly influences downstream applications such as epidemic spread, transportation demand, and infrastructure planning. Although recent work~\cite{yan2026guiding} has reported this phenomenon in specific leading models, its prevalence across different trajectory-generation paradigms remains unknown. We therefore pose \textbf{RQ1:} Does this displacement gap persist across diverse trajectory generation mechanisms, or is it restricted to specific model architectures?

Existing studies address this population-level mismatch through distinct analytical lenses. One line of work attributes population-level discrepancies in bottom-up generation to insufficient coordination among independently generated agents, which is associated with the synthetic population missing collective structures and emergent scaling regularities~\cite{yan2026guiding, park2023generative, liu2025human}. An alternative account traces the gap to missing activity-purpose semantics: trajectories are driven by underlying intents (home--work commuting, leisure, errands) that collectively shape decisions, and generators that ignore purpose tend to produce locally plausible but globally unstructured trips~\cite{bowman2001activity, shao2024chain, feng2020learning}. Human mobility studies demonstrate that aggregate displacement statistics are strongly shaped by heterogeneous individual behaviors and characteristic movement scales~\cite{han2011origin,yan2013diversity,alessandretti2020scales}, while trajectory-generation surveys consistently identify characteristic distances as vital spatial patterns for evaluating generative fidelity~\cite{luca2021survey}. Grounded in these insights, and rather than attempting to evaluate all potential mechanisms, we introduce and test a targeted, scale-based hypothesis: that the Population-Level Displacement Gap is strongly associated with systemic errors in how trips are allocated across distance scales. Under this scale-based lens, we anticipate different underlying failure modes across generation paradigms: for model-driven generators, rigid distance-kernel parameters may fail to capture multi-scale mixtures, leading to severe trip misallocation; conversely, for data-driven generators, scale-mixed objectives may allow dominant scales to monopolize the learning signal, smoothing away rare scales under finite model capacity. Consequently, we seek to answer \textbf{RQ2:} To what extent are population-level displacement errors associated with scale-level trip misallocation, and how does this misallocation differ between model-driven and data-driven generators?

This leads to \textbf{RQ3:} Can scale-decomposed generation mitigate the Population-Level Displacement Gap while preserving individual-level trajectory realism? To address this inquiry, we propose \textbf{ScaleTraj}, an autoregressive Transformer architecture for bottom-up trajectory generation that explicitly models movement scales. Existing autoregressive mobility generators often treat location prediction as sampling from a flattened spatial vocabulary, leaving scale structure implicit. Rather than predicting the next location from a monolithic spatial distribution, ScaleTraj decouples the process: it first predicts the target movement scale and subsequently models the destination cell conditioned on that scale. This factorization treats scale as an explicit routing variable while maintaining an individual-level generation paradigm.

In summary, we make three contributions. \textbf{Cross-Paradigm Diagnosis:} we evaluate representative model-driven and data-driven trajectory generators, spanning Markov, EPR, autoregressive recurrent, autoregressive Transformer, variational autoencoder (VAE), and generative adversarial network (GAN) paradigms, and show that the Population-Level Displacement Gap is prevalent across them. \textbf{Scale Fidelity Profile (SFP):} a scale-decomposed diagnostic metric that localizes displacement errors across spatial scales, exposing structural anomalies invisible to aggregate power-law parameters alone. \textbf{ScaleTraj:} an autoregressive Transformer with a Scale Selection Head and a Scale-Aware Cell Selection Head that mitigates the Population-Level Displacement Gap while preserving fidelity.
\section{Related Work}
\label{sec:related}

We position our contribution against two threads: the literature on multi-scale structure in human mobility, and the design space of bottom-up trajectory generators.

\subsection{Multi-Scale Structure in Human Mobility}
\label{sec:related-multiscale}

A central spatial regularity in human mobility is the distribution of travel distances $P(\Delta r)$, characterized by a heavy-tailed, truncated power law~\cite{gonzalez2008understanding,brockmann2006scaling}. Historically, this \emph{scale-free} property suggested that mobility lacked statistically distinct levels, such as ``neighborhood trips'' versus ``city trips.'' However, this invariance stands in tension with the hierarchical nature of urban structure and cognitive spatial representation~\cite{paasi2004place}. Subsequent research resolved this tension by demonstrating that individual mobility does not necessarily follow a power law~\cite{han2011origin}. Instead, aggregate scaling emerges as a result of mixing heterogeneous individual behaviors~\cite{yan2013diversity}. Alessandretti et al.~\cite{alessandretti2020scales} formalized this via container theory, proposing that individual mobility operates within nested spatial containers of characteristic sizes. The observed population-level power law is thus a superposition of these bounded, multi-level distributions.

This hierarchical perspective dictates a crucial constraint for the bottom-up trajectory generation explored in this paper: if the underlying data-generating process is multi-scale, a generator using a monolithic spatial distribution or a scale-agnostic objective may fail to replicate the population-level tail. Specifically, such models risk systemic scale-level trip misallocation, in which movements are incorrectly distributed across spatial containers—leading to the Population-Level Displacement Gap. While top-down models can bypass this by directly sampling from an aggregate power law, bottom-up generators should explicitly reconcile individual-level choices with these emergent scales to preserve aggregate-level realism after population aggregation.

\subsection{Trajectory Generation}

Existing trajectory generation approaches can be broadly categorized into model-driven and data-driven methods.
Model-driven generators encode mechanistic mobility rules, and two sub-families dominate. \textbf{Markov-chain models} represent trajectories as transitions over a discrete set of locations.
\textbf{EPR models} build on the Exploration and Preferential Return (EPR) framework~\cite{song2010modelling}, mixing uniform exploration with rank-weighted return.
TimeGeo~\cite{jiang2016timegeo} extends EPR with circadian patterns, activity-based scheduling, and rank-based destination selection, and is widely used for whole-day trip-chain generation.
Both sub-families offer interpretability, but their hand-crafted rules hard-code the scaling regularities they aim to describe~\cite{shinohara2025simple}: the scaling exponent is either an explicit input or an implicit consequence of the EPR kernel, and population-level accuracy can mask substantial individual-level deviations~\cite{napoli2024one}.

Data-driven generators learn trajectory distributions directly from data without explicit mobility rules.
Three generative families have been widely applied~\cite{luca2021survey}. \textbf{Autoregressive models} generate next locations conditioned on visit history.
Long short-term memory~\cite{10.1162/neco.1997.9.8.1735} (LSTM)-based generators factorize the trajectory likelihood as a product of conditional distributions over discrete locations and continuous sojourn times, autoregressively generating whole-day visit sequences.
TrajGPT~\cite{hsu2024trajgpt} recasts trajectory generation as a text-infilling problem: visit sequences are rearranged with special delimiters so that a Transformer can autoregressively fill variable-length gaps.
Recent work has also explored \textbf{large language model (LLM)-based} mobility generation. Rather than directly learning trajectory dynamics from mobility records alone, these approaches leverage LLMs to simulate mobility decisions, activity planning, or behavioral reasoning. For example, Chain-of-Planned-Behavior (CoPB)~\cite{shao2024chain} prompts LLMs with structured behavioral workflows to generate daily mobility trajectories, while subsequent studies have shown that LLM-generated populations may still exhibit discrepancies in aggregate mobility statistics after aggregation~\cite{yan2026guiding}. These findings suggest that improvements in semantic reasoning alone do not necessarily ensure fidelity to population-level mobility regularities.
\textbf{Variational autoencoders} compress trajectory into a continuous latent space.
TrajSynVAE~\cite{wang2023synthesizing} combines a variational autoencoder with a temporal point process: per-step latent codes capture user decision tendencies, and a deep network parameterizes the continuous-time probabilistic distribution over each visit's (location, timestamp, waiting-time) tuple.
\textbf{Adversarial methods} train a generator--discriminator pair to match the distribution.
MoveSim~\cite{feng2020learning} introduces a mobility-specific discriminator incorporating spatial and temporal regularizers.
STAR~\cite{wang2024spatiotemporal} augments a sequential generator with graph-structured branches encoding spatial adjacency and temporal periodicity.

Across those approaches, displacement is typically modeled through a single spatial distribution or generation objective. While these methods differ in architecture and inductive bias, explicit representations of the multi-scale hierarchical structure of human mobility remain largely unexplored. Moreover, existing evaluations primarily rely on aggregate mobility metrics, such as displacement distributions, radius of gyration, visitation frequency, and temporal regularity. Although informative, these metrics provide limited insight into how generation errors are distributed across spatial scales. Consequently, it remains unclear whether population-level displacement mismatches arise from broad distributional deviations or from systematic misallocation of trips across movement scales. This unresolved diagnostic gap motivates the central question of this paper: whether population-level displacement errors can be understood and mitigated through an explicit scale-based representation of human mobility.

\section{Preliminaries}
\label{sec:prelim}

\subsection{Definition}
A trajectory $\mathbf{X} = \{(t_i, c_i)\}_{i=1}^{L}$ is a time-ordered sequence of visits, where each $c_i$ indexes a tile in a uniform square tessellation over the study area~\cite{luca2021survey}. A globally defined $K$-scale partition $\{S_1, \ldots, S_K\}$ segments displacement space by distance thresholds into neighborhood / city / urban / region bands, operationalizing the hierarchical spatial structure described in container theory~\cite{alessandretti2020scales}. Whereas container theory allows the number and size of containers to vary
across individuals, we use a dataset-level set of ordered scale levels: evaluation assigns realized displacements to scale bands, while ScaleTraj uses the corresponding GMM components as routing variables for generation. Container theory posits that the population-level truncated power law emerges as a mixture of bounded-scale components:
\begin{equation}
P(\Delta r) = \sum_{k=1}^{K} \pi_k \cdot P(\Delta r \mid k),
\label{eq:superposition}
\end{equation}
where each $P(\Delta r \mid k)$ has bounded support within scale band $S_k$, and $\pi_k$ is the population-level mixing weight for scale $k$.
In the generative formulation of \S\ref{sec:method}, we denote the visit
timestamp by $a_i\equiv t_i$ and additionally model the stay duration
$\tau_i$.
\subsection{Datasets}
\label{sec:rq1-data}
Our evaluations are conducted on two real-world mobility datasets that differ in user population, sensing modality, and spatial resolution. \textbf{GeoLife}~\cite{zheng2010geolife} is a small-scale public GPS trajectory corpus, and \textbf{Tencent Beijing}~\cite{shao2024chain} is a large-scale mobility dataset. The two datasets were selected to represent different mobility-observation regimes. GeoLife contains high-resolution GPS trajectories from a relatively small user population, whereas Tencent Beijing is a large-scale released human-mobility corpus with coarser spatial resolution. Both datasets are snapped to a cell grid and split chronologically within each user into train and test partitions.

\subsection{Evaluation}
\label{sec:rq1-eval}
Population-level displacement fidelity is assessed using two fitted-parameter diagnostics, $|\Delta\beta|$ and $|\Delta\kappa|$ (km), obtained from the truncated power law~\cite{gonzalez2008understanding}, where $\beta$ is the scaling exponent and $\kappa$ the cutoff distance controlling tail truncation. These are reported together with the SFP, a scale-decomposed diagnostic metric defined in \S\ref{sec:rq2}. At the individual level, we report eight JSD-based metrics (all lower-is-better): step distance (SD) and radius of gyration (RG) for spatial fidelity; step interval (SI) and stay duration (STAY) for temporal fidelity; and daily travel distance (DTD), daily unique locations (DUL), visitation frequency (Visitation), and individual-rank correlation (IRank) for behavioral fidelity.

\section{RQ1: Evaluating the Displacement Gap Across Generative Paradigms}
\label{sec:rq1}

This section examines the extent to which individual-level trajectory generators replicate the emergent scaling properties of mobility data. We first verify that the displacement distributions of both datasets can be summarized by a truncated power law, then quantify the distribution-level discrepancies across all representative mechanisms under a unified evaluation framework.

\subsection{Cross-Paradigm Analysis}
\label{sec:rq1-results}

Prior work has noted discrepancies in emergent scaling for individual paradigms: Alessandretti et al.~\cite{alessandretti2020scales} for EPR-family models, Yan et al.~\cite{yan2026guiding} for LLM-based generators.
This analysis is extended systematically to representative major paradigms under a unified evaluation protocol.

\begin{figure*}[htbp]
\centering
\includegraphics[width=\textwidth]{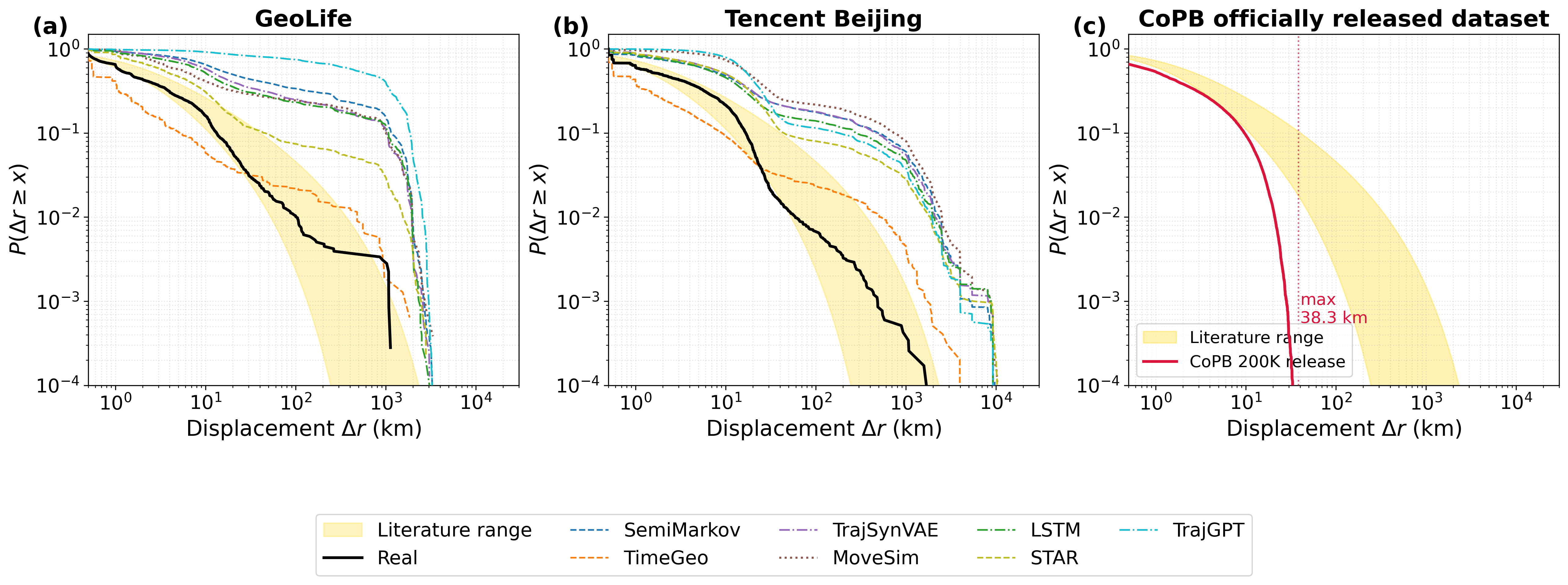}
\caption{Power-law diagnostic of trajectory generators. The vertical axis $P(\Delta r \ge x)$ gives the fraction of trips with displacement at least $x$. \textbf{(a, b)} Complementary cumulative distribution functions (CCDFs) of displacement distances on the GeoLife and Tencent Beijing test splits. The collected real-world data are shown in black, with the literature-consensus truncated-power-law family~\cite{gonzalez2008understanding} shown as external reference: the GeoLife curve lies largely within the band, whereas the Tencent Beijing curve approaches and falls below its lower edge in the tail. \textbf{(c)} Same diagnostic applied to CoPB's officially released dataset~\cite{shao2024chain}; its maximum observed displacement is $38.3$\,km, indicating a substantially shorter tail than the Tencent Beijing test distribution ($\kappa\!=\!129$\,km).}
\Description{Three log-log CCDF panels of displacement distance. Panels (a) and (b) show GeoLife and Tencent Beijing test-split real data as a black curve, with several colored baseline curves deviating from the band, mostly in the tail toward longer distances. Panel (c) shows the same diagnostic on the CoPB released dataset: the curve drops sharply to zero at 38.3 kilometers, far short of the gold band which extends past one thousand kilometers.}
\label{fig:powerlaw}
\end{figure*}

\begin{figure*}[htbp]
\centering
\includegraphics[width=\textwidth]{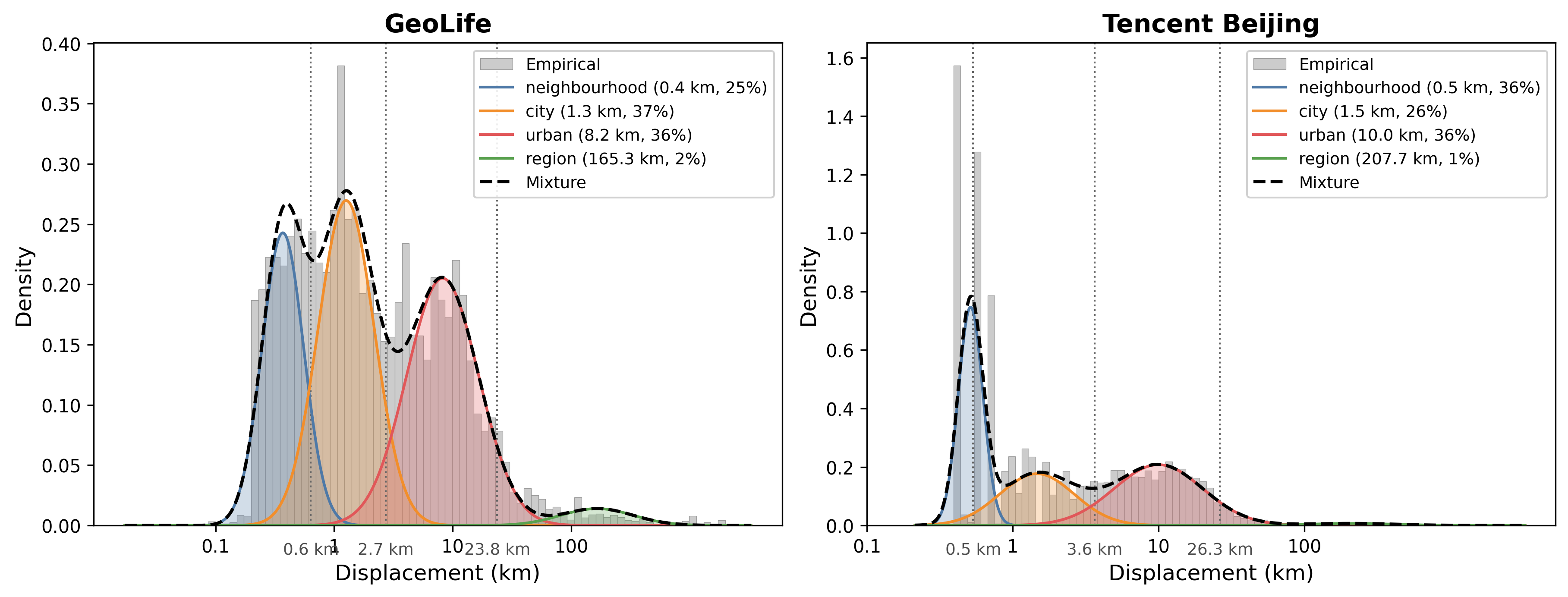}
\caption{Gaussian Mixture Model ($K=4$) fitted to log-displacement distances on the training set. Both datasets exhibit multi-modal structure in log-space, with four displacement bands from neighborhood to region. Vertical dotted lines indicate the dataset-specific scale boundaries used for the SFP diagnostic; the fitted GMM components provide the scale-conditioned distance structure used by ScaleTraj.}
\Description{Two side-by-side histograms of training log-displacement
distances, one for GeoLife and one for Tencent Beijing. Each histogram is overlaid with four fitted Gaussian components in different colors, and vertical dotted lines indicate the dataset-specific boundaries used to partition displacement into neighborhood, city, urban, and region bands
for evaluation.}
\label{fig:gmm}
\end{figure*}

As established in Table~\ref{tab:dataset-pl}, the displacement distributions on both datasets are well described by a truncated power law. The exponent $\beta$ is consistent across train and test ($|\Delta\beta|\!<\!.03$) on both datasets, while $\kappa$ shows greater sampling variation on the shorter test window, particularly on Tencent Beijing. For reference, prior work reports approximately $\beta\!\approx\!1.75\pm0.15$ and $\kappa\!\approx\!400$\,km~\cite{gonzalez2008understanding}; we use these literature-consensus values only as external context rather than as the evaluation target. Both datasets have displacement mass in all four Gaussian Mixture Model (GMM)-identified scale bands (Figure~\ref{fig:gmm}), and all reported $|\Delta\beta|$ and $|\Delta\kappa|$ values are computed against the corresponding test-split fits.

\begin{table}[t]
\centering
\caption{Truncated power-law fit ($\beta$, $\kappa$) for train and test splits. The exponent $\beta$ is consistent across splits ($|\Delta\beta|\!<\!.03$ on both datasets); $\kappa$ shows expected sampling variation on the shorter test window, especially on Tencent Beijing ($213\!\to\!129$\,km), but remains within the same order of magnitude.}
\label{tab:dataset-pl}
\small
\setlength{\tabcolsep}{5pt}
\begin{tabular}{llrrcc}
\toprule
Dataset & Split & Users & Days & $\beta$ & $\kappa$ (km) \\
\midrule
\multirow{2}{*}{GeoLife} & Train & 121 & 4{,}457 & 1.707 & 524 \\
                         & Test  & 121 & 1{,}171 & 1.730 & 356 \\
\midrule
\multirow{2}{*}{Tencent Beijing} & Train & 500 & 28{,}957 & 1.674 & 213 \\
                          & Test  & 500 &  7{,}507 & 1.686 & 129 \\
\bottomrule
\end{tabular}
\end{table}

To answer RQ1, we evaluate whether the Population-Level Displacement Gap persists across generation paradigms rather than being tied to a specific architecture. We assess seven representative generators: two model-driven
(SemiMarkov, TimeGeo) and five data-driven (LSTM, TrajSynVAE, MoveSim,
STAR, TrajGPT). Table~\ref{tab:emergence} summarizes the corresponding
fitted-parameter deviations for the seven controlled baselines. The SemiMarkov baseline uses the public implementation
released with TrajSynVAE~\cite{wang2023synthesizing}. We additionally examine the LLM-based generator CoPB~\cite{shao2024chain} as a separate case. Rather than retraining it, we diagnose the synthetic trajectory dataset the authors released alongside the method, which already exhibits a qualitatively different failure mode. Because this analysis uses the authors' released trajectories without
retraining under the unified protocol, CoPB is treated as a qualitative
external case and is not included in the controlled baseline comparison. As Figure~\ref{fig:powerlaw}(a,b) shows, every baseline deviates from
the test distribution, but the deviations fall into two opposite failure
modes.

The predominant mode is tail \emph{inflation}: most methods produce an inflated $\kappa$, over-representing long-distance trips. The magnitude varies widely across methods, and a close exponent match does not guarantee truncation fidelity: TimeGeo achieves the lowest $|\Delta\beta| = .077$ on GeoLife yet still inflates $\kappa$.

The opposite mode is tail \emph{collapse}, which appears in the examined LLM-based generator. Figure~\ref{fig:powerlaw}(c) applies the same diagnostic to CoPB's officially released synthetic dataset~\cite{shao2024chain}, a release accompanying the original method. Its maximum observed displacement is $38.3$\,km, indicating a substantially shorter tail than the Tencent Beijing test distribution, whose fitted cutoff is $\kappa\!=\!129$\,km. This sharply truncated support represents a qualitatively different failure mode from the tail inflation observed for most other generators.

Despite pointing in opposite directions, both modes share the same
underlying failure: no evaluated baseline simultaneously reproduces the
test-set scaling exponent \emph{and} its truncation structure. The Population-Level Displacement Gap is therefore consistently observed across all evaluated paradigms and is not restricted to any single architecture. These aggregate diagnostics, however, reveal only \emph{that} the distribution deviates, not \emph{where} in the displacement range the deviation concentrates.

\begin{table}[t]
\centering
\caption{Power-law diagnostics on the test set. $|\Delta\beta|$, $|\Delta\kappa|$ (km): deviation of fitted parameters from real data; lower is better.}
\label{tab:emergence}
\small
\setlength{\tabcolsep}{3.5pt}
\begin{tabular}{lcccc}
\toprule
 & \multicolumn{2}{c}{GeoLife} & \multicolumn{2}{c}{Tencent Beijing} \\
\cmidrule(lr){2-3}\cmidrule(lr){4-5}
Method & $|\Delta\beta|\!\downarrow$ & $|\Delta\kappa|\!\downarrow$ & $|\Delta\beta|\!\downarrow$ & $|\Delta\kappa|\!\downarrow$ \\
\midrule
SemiMarkov & .433 & 7{,}937  & .304 & 6{,}629 \\
TimeGeo    & .077 & 4{,}703  & .052 & 3{,}960 \\
LSTM       & .373 & 7{,}898  & .260 & 5{,}922 \\
TrajSynVAE & .382 & 6{,}566  & .296 & 5{,}967 \\
MoveSim    & .343 & 10{,}046 & .337 & 5{,}752 \\
STAR       & .191 & 5{,}267  & .221 & 3{,}366 \\
TrajGPT    & .536 & 13{,}816 & .280 & 2{,}422 \\
\bottomrule
\end{tabular}
\end{table}

\section{RQ2: Scale-Decomposed Diagnosis}
\label{sec:rq2}

RQ1 establishes the Population-Level Displacement Gap using aggregate displacement diagnostics. RQ2 asks where in the displacement range this discrepancy concentrates, and why. We proceed in four steps: (i) a training-set GMM defines statistically distinct displacement bands (\S\ref{sec:rq2-gmm}); (ii) the Scale Fidelity Profile (SFP) summarizes the discrepancy in how trips are allocated across these bands (\S\ref{sec:rq2-sfp}); (iii) applying SFP to the baselines localizes \emph{where} each one misallocates trips (\S\ref{sec:rq2-results}); and (iv) we explain \emph{why} these misallocations arise, with distinct mechanisms for model-driven and data-driven generators (\S\ref{sec:rq2-why}).

\subsection{Dataset Scale Bands}
\label{sec:rq2-gmm}
We partition displacement distances into $K{=}4$ bands by fitting a Gaussian Mixture Model to log-displacement on each dataset's training split. We adopt four ordered scale levels (neighborhood, city, urban, region), sorting the fitted components by increasing center. The default scale count is evaluated against alternative settings in Appendix~\ref{appendix:ksweep}.

The resulting bands are strongly imbalanced: the region band carries only
about $0.8\%$ of trips on Tencent Beijing and $3.8\%$ on GeoLife. This
imbalance is observed in both datasets, while the fitted components remain
distinct in log-distance space (Table~\ref{tab:gmm-params}). Preserving
this skewed allocation is exactly what SFP is designed to evaluate.

\subsection{Scale Fidelity Profile}
\label{sec:rq2-sfp}

Scale proportions are the quantity that matters for the Population-Level Displacement Gap. By the superposition structure in Equation~\eqref{eq:superposition}, the population-level displacement distribution is a mixture of per-scale components weighted by $\boldsymbol\pi$; if $\boldsymbol\pi$ is wrong, the aggregate distribution is wrong, regardless of how accurate each per-scale component is. For example, a generator that over-allocates trips to the region band will inflate the aggregate tail even if its intra-scale shapes are perfect; conversely, perfect intra-scale shapes cannot fix a misallocated $\boldsymbol\pi$. A useful diagnostic should therefore target the mixing weights directly, rather than the full displacement distribution, which also entangles grid resolution and intra-scale details that are less informative for our purposes.

This motivates comparing the real and generated scale-proportion vectors directly, where a lower discrepancy indicates closer scale-level fidelity.

\begin{definition}[Scale Fidelity Profile]
The Scale Fidelity Profile is defined as
\begin{equation}
\text{SFP} = \text{JSD}(\boldsymbol\pi^{\text{real}} \| \boldsymbol\pi^{\text{sim}}),
\label{eq:sfp}
\end{equation}
where $\boldsymbol\pi^{\text{real}}$ and $\boldsymbol\pi^{\text{sim}}$ are the scale-proportion vectors of the real and generated populations. Here $\pi_k$ is the fraction of step displacements assigned to band $S_k$, i.e.\ the count of displacements in $S_k$ over the total, where a step displacement is the haversine distance between two consecutive stay events in a trajectory. 
\end{definition}

\subsection{SFP Analysis of Baselines}
\label{sec:rq2-results}

\begin{figure*}[htbp]
\centering
\includegraphics[width=\textwidth]{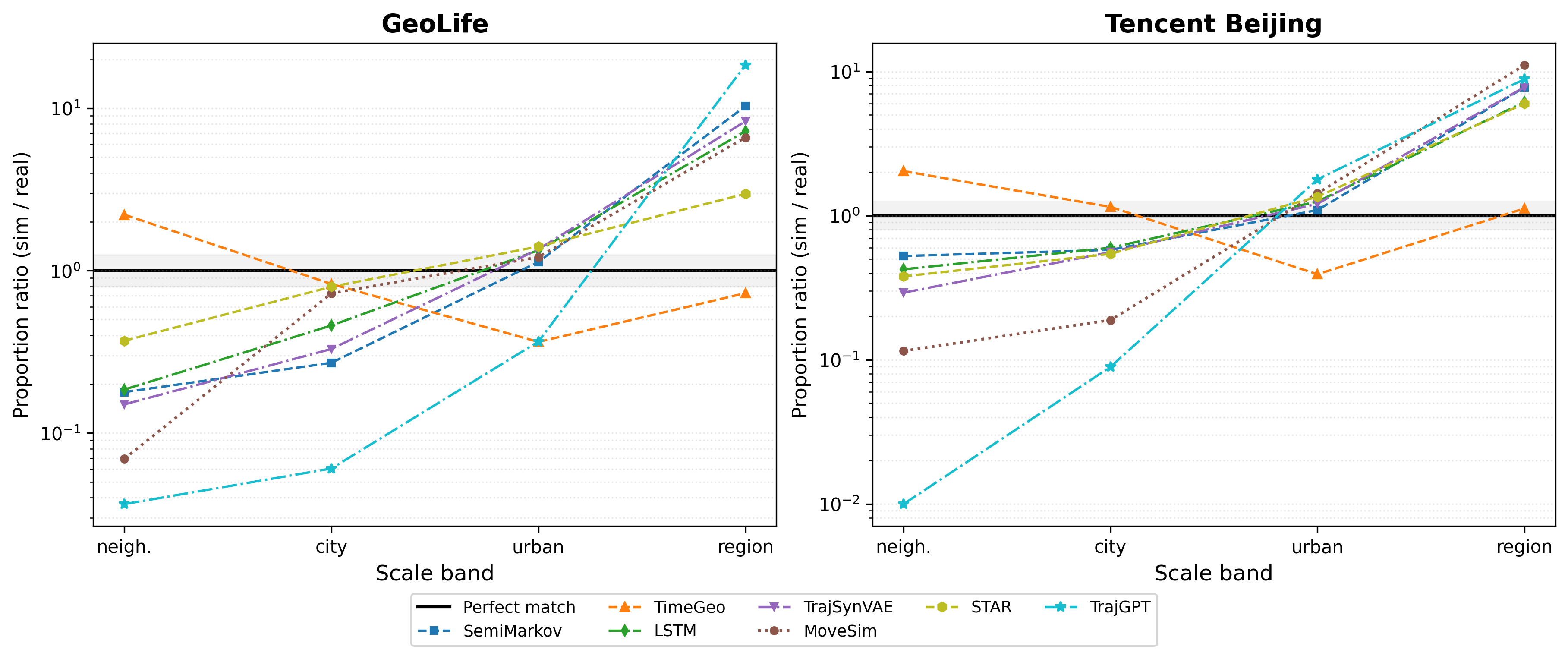}
\caption{Per-scale proportion ratio (sim/real) on log scale for seven baselines on the test set. The black line marks ratio $=1$, and the gray band spans $\pm 20\%$. Many methods show substantial scale-level misallocation, with short-range and long-range movements often deviating in different directions. These patterns expose compensating errors that are not visible from aggregate power-law parameters alone.}
\Description{Two log-scale line plots showing simulated-to-real
per-scale proportion ratios for the evaluated baselines across the neighborhood, city, urban, and region bands. A horizontal line marks perfect agreement and a gray band indicates a tolerance region. The profiles show substantial over- and under-allocation across scales.}
\label{fig:sfp-profile}
\end{figure*}

Figure~\ref{fig:sfp-profile} visualizes the per-scale proportion ratios (sim/real) for all baselines, with the corresponding SFP values reported in Table~\ref{tab:full}.

Model-driven generators exhibit distinct patterns of scale misallocation.
TimeGeo over-produces neighborhood-scale trips while under-producing
urban-scale trips, whereas SemiMarkov under-produces neighborhood-scale
trips and strongly over-produces region-scale trips. These contrasting
profiles yield different SFP values on GeoLife ($.057$ for TimeGeo versus
$.182$ for SemiMarkov), showing that scale-level errors can differ
substantially even within the same model family.

Data-driven generators share a more consistent direction of misallocation.
LSTM, TrajSynVAE, MoveSim, and STAR all over-produce region-scale trips
while under-producing neighborhood-scale trips, and TrajGPT shows a more
extreme variant of the same, reaching the highest SFP among all baselines
($.406$ on GeoLife). Moreover, similar aggregate scaling-exponent errors can mask substantially
different scale-level errors: on Tencent Beijing, TrajGPT and TrajSynVAE
have similar $|\Delta\beta|$ values ($.280$ versus $.296$), but markedly
different SFP scores ($.228$ versus $.080$).

Across both families, the dominant tendency is therefore to under-produce
neighborhood-scale trips and over-produce region-scale trips, with TimeGeo
as the notable exception. SFP spans a wide range (from $.038$ for STAR to
$.406$ for TrajGPT on GeoLife), showing that the Population-Level
Displacement Gap coincides with systematic scale-level misallocation.
Although this dominant pattern crosses generation paradigms, the mechanisms
underlying these scale-level errors may differ across model families, which
we examine next.

\subsection{Interpreting Scale Misallocation}
\label{sec:rq2-why}
The SFP diagnostic suggests that the Population-Level Displacement Gap is closely related to how generators allocate trips across spatial scales. The mechanisms differ between model-driven and data-driven generators.

For model-driven generators, the main limitation is that a small number of distance parameters is often used to represent a displacement distribution that is multi-scale in practice. For example, EPR-style models typically rely on a compact power-law jump kernel, while TimeGeo and related models add temporal structure but still use a limited distance-generation mechanism. Such models can fit an aggregate slope reasonably well, but they have limited control over how probability mass is distributed across neighborhood, city, urban, and region scales. As a result, a model may match a global power-law parameter while still over-producing trips at one scale and under-producing trips at another. This is illustrated by TimeGeo: it achieves the lowest $|\Delta\beta|$ among baselines on GeoLife, yet its SFP remains $.057$, showing that a close aggregate fit can coexist with substantial per-scale misallocation.

For data-driven generators, the issue is less about the number of hand-designed parameters and more about how the training signal is distributed. Standard autoregressive, VAE, and adversarial objectives are usually optimized over scale-mixed data, in which region-scale trips account for under $1\%$ of tokens on Tencent Beijing and under $4\%$ on GeoLife. Under unweighted token-level objectives, region-scale examples occur
roughly one to two orders of magnitude less frequently than
neighborhood-scale examples, depending on the dataset. This imbalance
can weaken the learning signal for rare scales. With finite model capacity, the network may preferentially fit the much
more frequent non-region scales, while sparse region-tail structure may be smoothed or absorbed into adjacent scales. Notably, this scarcity in training does not translate into scarcity at generation: because the rare scale is poorly modeled rather than simply suppressed, its probability mass is misplaced rather than removed, which is consistent with the inflated aggregate $\kappa$ observed for these generators in RQ1.

These observations do not prove that scale misallocation is the only source of the Population-Level Displacement Gap. Rather, they suggest why scale-level errors are plausible and why they appear across different generator families. Model-driven methods lack enough explicit degrees of freedom to control
scale proportions, while data-driven methods receive weaker training signals for rare scales because of imbalanced token frequencies. That these distinct mechanisms can produce the same symptom points to a
shared omission rather than a paradigm-specific flaw: neither family represents movement scale as an explicit, separately controllable quantity. This shared omission directly motivates the design of ScaleTraj, which promotes scale from an implicit by-product of the spatial softmax to an explicit routing variable, giving each scale band dedicated capacity and a controllable mixing weight.

\section{RQ3: ScaleTraj Framework}
\label{sec:method}

\begin{figure*}[t]
\centering
\includegraphics[width=\textwidth, trim={0 150 0 0}, clip]{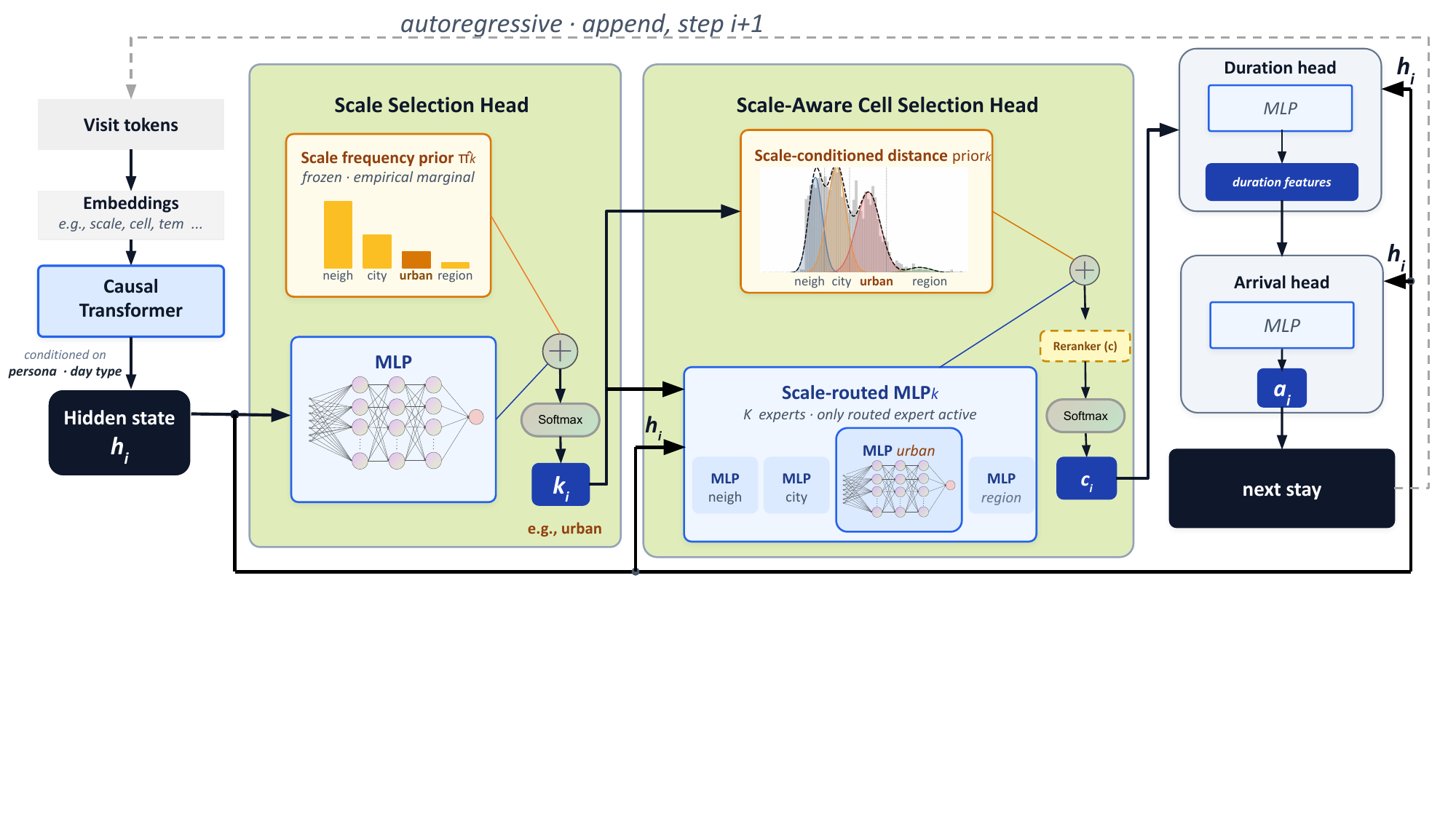}
\caption{ScaleTraj architecture.
Past visits are embedded and encoded by a causal Transformer ($L{=}4$, $H{=}4$, $d{=}128$) into $h_i$. The \textbf{Scale Selection Head} combines a frozen training-set prior $\hat{\pi}_k$ with a learned context MLP to sample $k_i$ (here ``urban'').
The \textbf{Scale-Aware Cell Selection Head} routes the current model state
through a scale-specific MLP and combines it with a scale-conditioned
Gaussian distance prior; an optional composite inference-time reranking
stage $\mathrm{Reranker}(c)$ is applied before sampling $c_i$. Duration and arrival heads complete the visit, and $(k_i, c_i, \tau_i, a_i)$
is fed back autoregressively. \textbf{Colors:} orange = frozen; blue = learned.}
\Description{A flow diagram of the ScaleTraj architecture. Past visit tokens are embedded and passed through a causal Transformer encoder into a hidden state. The hidden state feeds a Scale Selection Head, shown combining a frozen training-set scale-frequency prior with a learned MLP to sample a scale band, and a Scale-Aware Cell Selection Head, shown routing the hidden state through one of four per-scale MLPs and adding a scale-conditioned Gaussian distance prior plus an optional inference-time reranker before sampling a destination cell. The duration head predicts duration outputs and intermediate duration features, which are also provided to the arrival head. The sampled scale, cell, duration, and arrival are fed back into the sequence for the next autoregressive step. Frozen components are colored orange and learned components are colored blue.}
\label{fig:framework}
\end{figure*}

RQ2 localized the Population-Level Displacement Gap to systematic misallocation of trips across scale bands. ScaleTraj addresses this by treating scale as an explicit routing variable rather than an emergent property of a single spatial softmax. Figure~\ref{fig:framework}
summarizes the overall architecture. We first present the scale-decomposed factorization (\S\ref{sec:scale-boundaries}), then the Transformer backbone producing the hidden state $h_i$ (\S\ref{sec:backbone}), and finally the prediction heads that implement the factorization (\S\ref{sec:heads}).

\subsection{Scale-Decomposed Factorization}
\label{sec:scale-boundaries}

Standard autoregressive trajectory generators predict cell $c_i$ from a single softmax over the full cell vocabulary, implicitly marginalizing over scales. As described in the previous section, this formulation allocates capacity in proportion to token frequency, starving rare region-scale trips. Rather than letting rare region-scale trips compete with dominant neighborhood-scale trips in a single softmax, ScaleTraj first decides the scale $k$ and then predicts the cell $c$ conditioned on $k$.

Concretely, ScaleTraj decomposes the prediction into a cascade of conditionals. At each step $i$, conditioned on the Transformer hidden state $h_i$:
\begin{equation}
P(\text{traj}) = \prod_i
P(k_i \mid h_i)\,
P(c_i \mid k_i,h_i)\,
P(\tau_i \mid k_i,c_i,h_i)\,
P(a_i \mid k_i,c_i,h_i).
\label{eq:factor}
\end{equation}
where $k_i \in \{1,\ldots,K\}$ is the spatial scale, $c_i$ is the discrete grid cell, $\tau_i$ is the stay duration, and $a_i$ is the arrival time. Scale is predicted first by the \emph{Scale Selection Head}; cell prediction is then explicitly conditioned on scale by the \emph{Scale-Aware Cell Selection Head}, so each scale band receives dedicated model capacity regardless of its population frequency.

ScaleTraj uses the training-set GMM introduced in \S\ref{sec:rq2-gmm} to define its ordered scale-conditioned components. The selected $k_i$ acts as a routing variable for destination prediction rather than as a hard constraint on the realized displacement. Because the GMM is re-fit from each dataset's own displacement
distribution rather than using fixed scale components across datasets, the same architecture can adapt its scale-conditioned components to dataset-specific displacement structure.

\subsection{Transformer Backbone}
\label{sec:backbone}
A causal Transformer~\cite{vaswani2017attention} serves as the autoregressive backbone because daily mobility decisions depend on non-local visit history: a return trip is conditioned on where the day began, and an evening destination on the accumulated trajectory rather than only the previous stop. Within the supplied context window, self-attention allows each prediction to condition directly on any earlier visit through explicit pairwise attention. This is useful for scale decomposition in particular, since the appropriate movement scale (e.g., a long commute home versus a short errand) is often determined by visits several steps earlier.

Each visit token concatenates a cell embedding, a scale embedding, Fourier-encoded temporal features, an anchor-conditioned embedding carrying home/work cells and a user-latent code, and persona and time-budget features; Rotary Position Embedding (RoPE)~\cite{su2021roformer} encodes visit order and a causal mask restricts $h_i$ to visits $1,\ldots,i{-}1$. We prepend up to $C{=}3$ previous days of the same user's stays as history context. The four prediction heads branch from $h_i$ and are trained jointly with teacher forcing plus scheduled sampling on the scale step. More details are in Appendix~\ref{appendix:hparams}.

\subsection{Prediction Heads}
\label{sec:heads}

The two scale-aware heads are the primary architectural contribution of this paper. They realize the factorization of Eq.~\eqref{eq:factor}: the \emph{Scale Selection Head} produces $P(k_i \mid h_i)$ and the \emph{Scale-Aware Cell Selection Head} produces $P(c_i \mid k_i, h_i)$.

\paragraph{Scale Selection Head.}
The Scale Selection Head produces a $K$-dimensional logit vector:
\begin{equation}
\operatorname{logit}_i(k)
=
\alpha_m \log \hat{\pi}_k
+
\alpha_h
\bigl[\operatorname{MLP}_{\mathrm{scale}}(h_i)\bigr]_k .
\end{equation}
where $\hat{\pi}_k$ is the marginal scale proportion estimated from the
training set, $\alpha_m$ and $\alpha_h$ are fixed hyperparameters, and
$\operatorname{MLP}_{\mathrm{scale}}$ is a two-layer MLP whose output
provides a context-dependent residual adjustment to the marginal-prior
logits. Keeping $\alpha_m$ and $\alpha_h$ fixed retains an explicit prior
contribution and is intended to discourage collapse toward a
frequency-dominated free softmax.

\paragraph{Scale-Aware Cell Selection Head.}
Once $k_i$ is determined, cell logits combine two additive terms:
\begin{equation}
\label{eq:cell}
\ell_i(c)
=
\alpha_d\bigl[g_{k_i}(c)-J_{k_i}(c)\bigr]
+
\alpha_r\bigl[r^{(k_i)}(h_i)\bigr]_c .
\end{equation}
where $g_{k_i}(c)$ is a GMM-based Gaussian distance prior,
given by the Gaussian log-density of
$\log d(c_{i-1}, c)$ under scale $k_i$'s fitted component, $r^{(k_i)}$ is the scale-specific residual MLP, $J_{k_i}(c)$ is a correction term for the non-uniform density of candidate cells, and $\alpha_d, \alpha_r$ are fixed coefficients. $P(c_i \mid k_i, h_i) = \text{softmax}(\ell)$. The prior $g_{k_i}$ adds a quadratic log-penalty growing with distance from the scale's characteristic range, rather than relying on the model to learn spatial boundaries from data alone.

The residual term $r^{(k_i)}$ is realized as $K$ independent two-layer MLPs, one per scale level. For compactness, the notation
$r^{(k_i)}(h_i)$ suppresses two auxiliary inputs used in the implementation: the selected-scale feature and the previous-cell embedding. We retain the name ``residual MLP'' because $r^{(k_i)}$ acts as a learnable residual correction added to the geometric prior $g_{k_i}$, not because the MLP itself contains a skip connection. At each training step, only the routed scale-specific expert receives
gradients; routing uses the ground-truth scale under teacher forcing and the predicted scale under scheduled sampling. This gives each scale dedicated capacity rather than forcing all movements through a single shared residual predictor. The ablation verifies that replacing a single shared MLP with these per-scale MLPs is associated with improved heavy-tail fidelity. Additional details on the correction term and persona-conditioned user
latents are provided in Appendix~\ref{appendix:arch}.

\paragraph{Duration and arrival heads.}
Stay duration $\tau_i$ is predicted by a two-layer MLP from $h_i$, $e_{k_i}$, and $e_{c_i}$, output over non-uniform duration bins (denser at short durations to match empirical heavy-tailed stays); an auxiliary Gaussian regression head $(\mu, \log\sigma)$ is combined at weight $\lambda_g{=}0.5$. Arrival time $a_i$ is predicted from the same inputs, augmented with the duration head's predicted Gaussian parameters $(\mu_i,\log\sigma_i)$.

\subsection{Training and Generation}
\label{sec:rq3-gen}

Training minimizes categorical cross-entropy losses on the cascade outputs,
together with a Gaussian duration negative log-likelihood and auxiliary end-of-sequence (EOS), activity-type, day-level statistics, and persona-conditioned user-latent regularization terms.

Generation samples autoregressively at frozen parameters. The cell head is sharpened by a fixed temperature and augmented with two inference-time rerankers motivated by well-documented mobility regularities: a global-frequency term biasing generation toward popular places, and a revisit reward encouraging the routine return-trips that dominate individual mobility~\cite{song2010modelling}. Causal masks enforce temporal validity, and generation terminates at the EOS threshold or end-of-day. More details are in Appendix~\ref{appendix:hparams}.
\section{Experiments}
\label{sec:experiments}
\begin{table*}[t]
\centering
\caption{Full evaluation on both datasets. All metrics $\downarrow$ (lower is better); per-dataset best in bold.}
\label{tab:full}
\footnotesize
\setlength{\tabcolsep}{3.5pt}
\begin{tabular}{l*{11}{c}|*{11}{c}}
\toprule
 & \multicolumn{11}{c|}{Tencent Beijing} & \multicolumn{11}{c}{GeoLife} \\
\cmidrule(lr){2-12}\cmidrule(lr){13-23}
Method
  & SD & RG & SI & STAY & DTD & DUL & Visit & IRank & SFP & $|\Delta\beta|$ & $|\Delta\kappa|$
  & SD & RG & SI & STAY & DTD & DUL & Visit & IRank & SFP & $|\Delta\beta|$ & $|\Delta\kappa|$ \\
\midrule
SemiMarkov  & .089 & .199 & .282 & .101 & .405 & .487 & .319 & .440 & .067 & .304 & 6{,}629
            & .234 & .351 & .294 & .142 & .575 & .572 & .387 & .586 & .182 & .433 & 7{,}937 \\
TimeGeo     & .056 & .143 & .032 & .110 & .127 & .037 & .269 & .039 & .040 & .052 & 3{,}960
            & .158 & .219 & .111 & .174 & .207 & .284 & .435 & .136 & .057 & .077 & 4{,}703 \\
LSTM        & .081 & .197 & .182 & .106 & .467 & .548 & .318 & .502 & .057 & .260 & 5{,}922
            & .174 & .202 & \textbf{.014} & \textbf{.030} & .327 & .227 & .118 & .264 & .121 & .373 & 7{,}898 \\
TrajSynVAE  & .106 & .169 & .088 & \textbf{.021} & .284 & .256 & .112 & .234 & .080 & .296 & 5{,}967
            & .204 & .251 & .020 & .038 & .345 & .160 & .070 & .207 & .155 & .382 & 6{,}566 \\
MoveSim     & .231 & .262 & .178 & .091 & .318 & .210 & .158 & .258 & .186 & .337 & 5{,}752
            & .197 & .129 & .122 & .123 & .140 & .057 & .143 & .061 & .117 & .343 & 10{,}046 \\
STAR        & .093 & .147 & .297 & .274 & .186 & .294 & .175 & .161 & .064 & .221 & 3{,}366
            & \textbf{.099} & \textbf{.075} & .353 & .363 & .096 & .087 & .087 & .086 & .038 & .191 & 5{,}267 \\
TrajGPT     & .272 & .432 & .212 & .267 & .570 & .575 & .212 & .587 & .228 & .280 & 2{,}422
            & .459 & .672 & .200 & .097 & .693 & .693 & .399 & .693 & .406 & .536 & 13{,}816 \\
\textbf{ScaleTraj} & \textbf{.039} & \textbf{.023} & \textbf{.007} & .055 & \textbf{.016} & \textbf{.018} & \textbf{.018} & \textbf{.019} & \textbf{.003} & \textbf{.021} & \textbf{24}
            & .123 & .080 & .047 & .032 & \textbf{.061} & \textbf{.054} & \textbf{.038} & \textbf{.021} & \textbf{.003} & \textbf{.026} & \textbf{131} \\
\bottomrule
\end{tabular}
\end{table*}

This section evaluates ScaleTraj against the baselines under the unified protocol of \S\ref{sec:rq1-eval}. Synthetic generation is aligned to the held-out test user-day records. ScaleTraj is trained from scratch on a single A6000 GPU. Beyond the main comparison, an ablation over the two cell-head knobs (\S\ref{sec:ablation}) and a sweep over the number of scale bands $K$ is reported in Appendix~\ref{appendix:ksweep}. Key configuration details are reported in Appendix~\ref{appendix:hparams}.

\subsection{ScaleTraj Results}
\begin{figure}[htbp]
\centering
\includegraphics[width=\columnwidth]{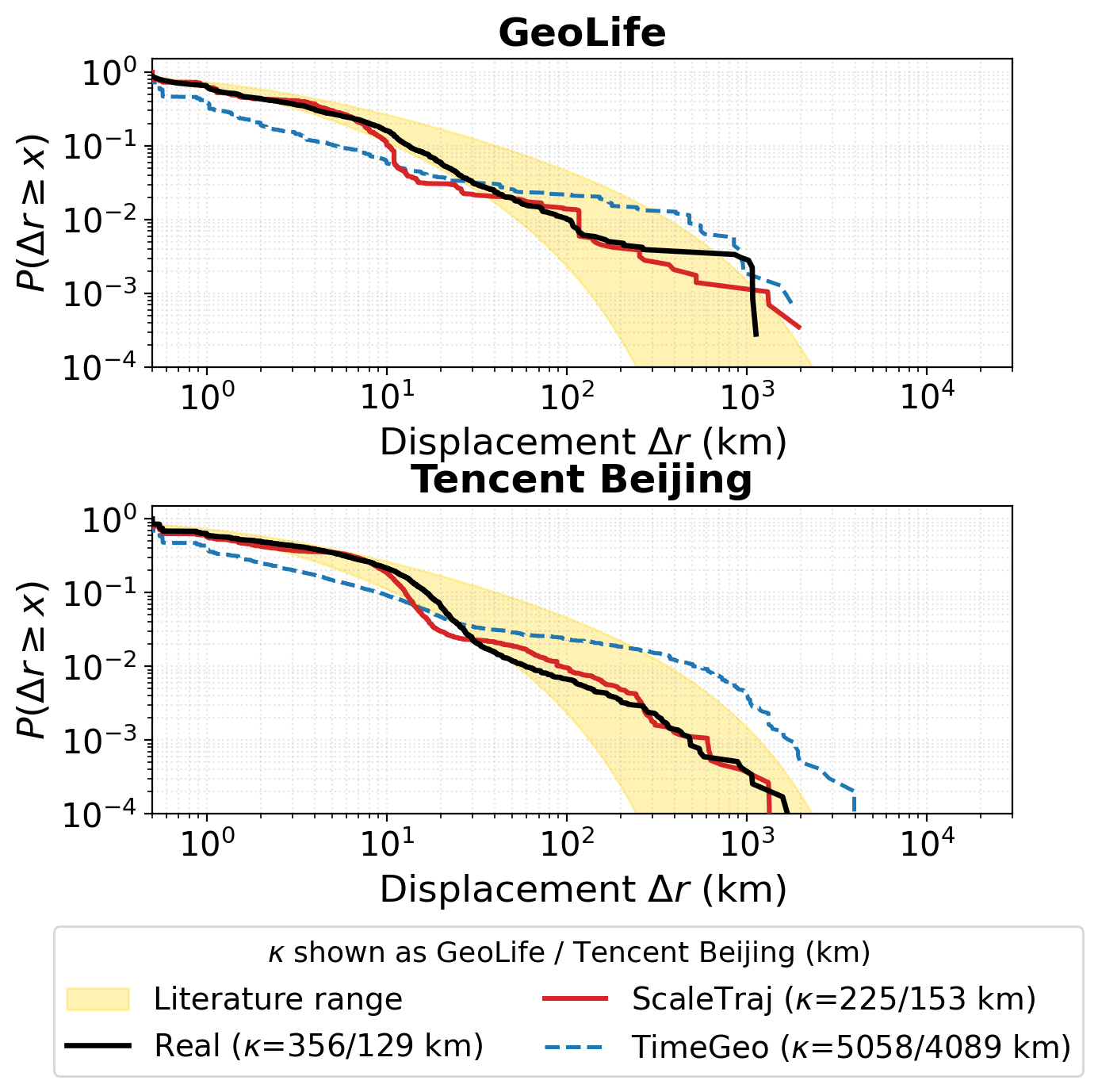}
\caption{Displacement CCDF comparison of ScaleTraj (red), the real
test-set distribution (black), and TimeGeo (blue dashed), the strongest
baseline in terms of $|\Delta\beta|$. The literature-consensus band is
shown as external reference.}
\Description{Two log-log CCDF plots, one for GeoLife and one for Tencent
Beijing. Each panel compares the real test-set distribution, ScaleTraj,
and TimeGeo, with the literature-consensus band shown as external
reference. ScaleTraj more closely follows the real-data CCDF, while
TimeGeo exhibits a larger deviation in the tail.}
\label{fig:scaletraj_ccdf}
\end{figure}

The primary evaluation considers the three diagnostics directly tied to
RQ1 and RQ2: SFP, $|\Delta\beta|$, and $|\Delta\kappa|$. As shown in
Table~\ref{tab:full}, ScaleTraj achieves the best performance on all three
diagnostics on both datasets. This indicates that the improvement is
observed at both levels targeted by the proposed framework: the allocation
of trips across spatial scales and the resulting aggregate displacement
distribution.

ScaleTraj models displacement through scale-specific components rather than relying on a single scale-agnostic
spatial prediction process. The Scale Selection Head explicitly controls
the allocation of movements across scale bands, while the Scale-Aware Cell
Selection Head models destination choice conditioned on the selected scale.
The improvements in both SFP and the fitted power-law diagnostics in
Table~\ref{tab:full} are consistent with this decomposition: better
preservation of scale proportions is accompanied by a closer match to the
aggregate test-set displacement distribution.

Figure~\ref{fig:scaletraj_ccdf} provides the corresponding distributional
comparison. ScaleTraj closely follows the test-set CCDF across the
displacement range, while TimeGeo, the strongest baseline in terms of
$|\Delta\beta|$, still exhibits a substantially different tail. This
comparison illustrates why matching the aggregate scaling exponent alone
is insufficient for displacement fidelity.

Individual-level metrics provide an additional preservation check. As
reported in Table~\ref{tab:full}, ScaleTraj remains competitive across the
spatial, temporal, and behavioral metrics while substantially improving
the scale-level and aggregate-displacement diagnostics. Its strongest
individual-level results occur on the behavioral metrics, while several
baselines remain better on selected spatial or temporal measures. Overall,
the results indicate that improved population-level displacement fidelity
does not require a broad sacrifice in individual-level trajectory
fidelity.

\subsection{Ablation}
\label{sec:ablation}

The cell head has two architectural knobs: the number of scale bands $K$ and whether the residual MLP is shared across scales or split into $K$ per-scale MLPs (Table~\ref{tab:ablation}).

\begin{table}[htbp]
\centering
\caption{Ablation on the cell head's two knobs: $K$ and residual-MLP sharing. Bottom row is the full ScaleTraj system.}
\label{tab:ablation}
\small
\setlength{\tabcolsep}{3pt}
\begin{tabular}{lcccccc}
\toprule
 & \multicolumn{3}{c}{Tencent Beijing} & \multicolumn{3}{c}{GeoLife} \\
\cmidrule(lr){2-4}\cmidrule(lr){5-7}
Variant & SFP$\downarrow$ & $|\Delta\kappa|\!\downarrow$ & $|\Delta\beta|\!\downarrow$ & SFP$\downarrow$ & $|\Delta\kappa|\!\downarrow$ & $|\Delta\beta|\!\downarrow$ \\
\midrule
$K\!=\!1$, shared MLP                & .038 & 74 & .169 & .043 & 343 & .869 \\
$K\!=\!4$, shared MLP                & .009 & 58 & .050 & .002 & 19{,}644 & .121 \\
ScaleTraj & .003 & 24 & .021 & .003 &    131 & .026 \\
\bottomrule
\end{tabular}
\end{table}

\paragraph{$K\!=\!1$, shared MLP (no scale decomposition).}
With $K\!=\!1$ the scale-conditioned Gaussian collapses to a single global Gaussian fit on the full displacement distribution, and the per-scale MLP set collapses to one shared MLP. This single-scale baseline leaves substantial residual error on every diagnostic ($|\Delta\beta|\!=\!.169$/$|\Delta\kappa|\!=\!74$ on Tencent Beijing and $|\Delta\beta|\!=\!.869$/$|\Delta\kappa|\!=\!343$ on GeoLife). A single Gaussian cannot represent the multi-scale dataset's displacement distribution, so it concentrates displacement mass on short-range cells; the small Tencent Beijing $|\Delta\kappa|\!=\!74$ is a maximum likelihood estimation (MLE) artifact of that collapsed support, not evidence of good tail fidelity. $|\Delta\kappa|$ becomes informative only once a generator has the right qualitative shape.

\paragraph{$K\!=\!1 \to K\!=\!4$: scale decomposition improves scale fidelity.}
Expanding to four scale bands (one global Gaussian becomes four scale-conditioned Gaussians, and the Scale Selection Head chooses among them), with the shared MLP held fixed, drops SFP from $.038$ to $.009$ on Tencent Beijing and from $.043$ to $.002$ on GeoLife; aggregate $|\Delta\beta|$ drops from $.169$ to $.050$ on Tencent Beijing and from $.869$ to $.121$ on GeoLife. Scale decomposition alone reduces the population-level scale misallocation. On GeoLife $|\Delta\kappa|$ temporarily inflates to $19{,}644$ before
per-scale MLPs resolve it, suggesting that the shared MLP remains a bottleneck.

\paragraph{Shared MLP $\to K\!=\!4$ per-scale MLPs: per-scale capacity improves heavy-tail fidelity.}
A shared MLP must model all scales jointly despite their highly imbalanced token frequencies; on GeoLife, this configuration exhibits
$|\Delta\kappa|\!=\!19{,}644$. Splitting the residual into $K\!=\!4$ per-scale MLPs
(with only the routed scale-specific MLP active at each training step) drops $|\Delta\kappa|$ to $131$ on GeoLife and to $24$ on Tencent Beijing; $|\Delta\beta|$ also drops to $.021$/$.026$. Per-scale MLPs strictly dominate the shared MLP on both power-law diagnostics, with only a marginal SFP cost on GeoLife. We adopt the $K\!=\!4$ per-scale configuration as the released ScaleTraj.

\section{Discussion and Limitations}
\label{sec:discussion}
This work's evaluation spans representative model-driven and data-driven generators, providing evidence that the Population-Level Displacement Gap is not confined to a single model. The evaluated set, nevertheless, is not exhaustive; diffusion-based, hybrid-based, and other emerging mobility generators are not included here and remain to be systematically examined. Whether the Population-Level Displacement Gap extends to these paradigms, and whether scale decomposition offers similar benefits there, is an open question.

Generalization across datasets also warrants further study. Both datasets
are collected in China, and GeoLife contains a relatively small number of users, which limits
the diversity of geographic settings and population structures represented
in the current evaluation. Mobility patterns may differ across cities and
regions because of variations in urban form, transportation infrastructure, and observation scale, potentially affecting both the learned scale
boundaries and the resulting displacement distributions. In addition, displacement statistics can be sensitive to stay-point extraction, spatial
discretization, and other trajectory preprocessing choices, particularly
at neighborhood and city scales. Future work should therefore examine
whether the observed scale-level misallocation patterns remain stable across
additional generators, geographic regions, population sizes, and preprocessing configurations. Such evaluations would help determine which
aspects of the observed scale structure are broadly transferable and which
are dataset-specific.

The truncated power law and SFP are diagnostics of distributional fidelity, not complete definitions of mobility realism. Real displacement patterns are shaped by transportation infrastructure, land use, culture, and observation timescale, and deviations in $\beta$, $\kappa$, or scale proportions do not by themselves imply behaviorally unrealistic trajectories. SFP should therefore be viewed as complementary to conventional spatial, temporal, and behavioral metrics rather than as their replacement. Moreover, the absolute SFP values depend on implementation choices such as the smoothing constant and logarithm base, so SFP is best interpreted as a relative ranking across methods under a fixed evaluation protocol rather than as an absolute measure.

ScaleTraj explicitly incorporates training-set scale statistics via GMM-derived bands and priors. Its gains therefore show the benefit of preserving scale structure rather than discovering it without prior information. A natural concern is that ScaleTraj encodes the same scale statistics on which it is evaluated. However, the reported individual-level JSD metrics are not directly
optimized, yet ScaleTraj improves on most of them; the gains therefore extend beyond the scale statistics it explicitly incorporates. The individual-level metrics indicate no broad loss of trajectory realism, but do not establish improvements in all behavioral dimensions. Our use of learned user latent codes also limits the current evaluation to known-user generation; unseen-user generalization remains future work. Finally, our results come from a single generation run per configuration; reporting variance across repeated generation runs is left to future work.
\section{Conclusion}
\label{sec:conclusion}
The Population-Level Displacement Gap, a systematic failure of bottom-up trajectory generators to reproduce the displacement distribution after aggregation, is shown to be prevalent across representative model-driven and data-driven paradigms. To diagnose the gap, the Scale Fidelity Profile reveals systematic
misallocation of trips across spatial scales. Building on this diagnosis, ScaleTraj treats scale as an explicit routing variable and substantially reduces the gap while preserving individual-level fidelity. These results indicate that the Population-Level Displacement Gap is a diagnosable, scale-localized phenomenon, and that scale-aware evaluation offers a useful lens for future mobility generators.

\begin{acks}
This research was funded by the U.S. National Science Foundation under Grant 2330504. The views and conclusions presented in this study are those of the authors and do not necessarily reflect the official policy or position of NSF. 

The authors disclose use of generative AI tools (Claude and ChatGPT) to assist with manuscript editing, code review, experiments, and analyses; all research design and interpretation of results were performed by the authors.
\end{acks}

\section*{Ethics and Privacy Statement}
Both datasets used in this work are publicly released research corpora used under their intended terms. Synthetic trajectory generation carries dual-use considerations. On the benefit side, it can reduce the need to share raw mobility traces while supporting
downstream research such as epidemic simulation and EV charging planning. On the risk side, mobility generators could in principle be misused to profile movement patterns or to fabricate plausible traces for surveillance. We note that ScaleTraj is currently evaluated in a known-user setting, replaying trained user codes so that generated and real populations share the same anchors; it is therefore a tool for augmenting and studying existing populations rather than for reconstructing traces of unseen individuals. All experiments use only de-identified data, and we release our code to support reproducibility and to encourage downstream use under appropriate consent and governance.

\bibliographystyle{ACM-Reference-Format}
\bibliography{references}

\appendix

\section{Hyperparameters}
\label{appendix:hparams}

Unless noted otherwise, all values below are shared across both datasets and across every reported configuration (the three ablation rows of \S\ref{sec:ablation} and the $K$-sweep of Appendix~\ref{appendix:ksweep}); only the number of scale bands $K$ and the residual-MLP sharing flag vary between configurations. The Transformer backbone uses hidden size $d{=}128$, $L{=}4$ layers, and $H{=}4$ attention heads. Optimization uses AdamW with learning rate $3\!\times\!10^{-4}$ and batch size $64$, with validation-based early stopping (no curriculum or pre-training). The cell head combines the GMM-based distance prior with the per-scale residual MLP via fixed coefficients $\alpha_d{=}2.0$ and $\alpha_r{=}0.5$ (Eq.~\eqref{eq:cell}); the scale head's two fixed coefficients are $\alpha_m{=}\alpha_h{=}1.0$. Inference applies a cell-head temperature $T_c{=}0.3$ together with the two additive rerankers of \S\ref{sec:rq3-gen} at $\alpha_{\text{freq}}{=}\alpha_{\text{ret}}{=}0.3$. Two inference-time termination parameters are dataset-specific: the EOS threshold $\theta_{\text{EOS}}$ is $0.5$ on Tencent Beijing and $0.7$ on GeoLife, and the minimum-stops floor is $1$ on Tencent Beijing and $2$ on GeoLife (reflecting each corpus's empirical day-length distribution). The remaining auxiliary-loss weights and tokenization settings are fixed across datasets and configurations; their exact values are specified in the released training and reproduction code.

\section{Sensitivity to the number of scale bands \texorpdfstring{$K$}{K}}
\label{appendix:ksweep}

\subsection{Fitted \texorpdfstring{$K{=}4$}{K=4} GMM components}
For reference, Table~\ref{tab:gmm-params} reports the full $K{=}4$ GMM fit
used throughout the paper. The fitted components capture different regions
of the log-displacement distribution, with the region component carrying
relatively small mixture weight at this observation timescale.

\subsection{\texorpdfstring{$K$}{K}-sweep on the ScaleTraj configuration}
To verify that the default $K\!=\!4$ choice is not narrowly tuned, we additionally train the ScaleTraj configuration (per-scale MLPs of Table~\ref{tab:ablation}) at $K\!\in\!\{5,6\}$ on both datasets; only $K$ varies, with the GMM refit at the new number of components. Table~\ref{tab:ksweep} reports the three scale-emergence metrics.

\begin{table}[t]
\centering
\caption{Fitted $K{=}4$ GMM components on log-displacement: per-component center (km), log-space standard deviation $\sigma_{\log}$, mixture weight, and per-band token share after boundary
assignment.}
\label{tab:gmm-params}
\small
\setlength{\tabcolsep}{4pt}
\begin{tabular}{llrrrr}
\toprule
Dataset & Band & $e^{\mu_{\log}}$ (km) & $\sigma_{\log}$ & weight & tokens \\
\midrule
\multirow{4}{*}{GeoLife}
  & neighborhood    &   0.37 & 0.41 & 0.247 & 45.6\% \\
  & city             &   1.26 & 0.55 & 0.368 & 24.3\% \\
  & urban            &   8.23 & 0.70 & 0.360 & 26.3\% \\
  & region           & 165.27 & 0.70 & 0.025 &  3.8\% \\
\midrule
\multirow{4}{*}{Tencent Beijing}
  & neighborhood    &   0.51 & 0.19 & 0.362 & 53.3\% \\
  & city             &   1.46 & 0.59 & 0.261 & 20.5\% \\
  & urban            &   9.97 & 0.70 & 0.365 & 25.4\% \\
  & region           & 207.75 & 0.70 & 0.012 &  0.8\% \\
\bottomrule
\end{tabular}
\end{table}

\begin{table}[htbp]
\centering
\caption{$K$-sweep on the ScaleTraj configuration; lower is better. $K{=}4$ is the default.}
\label{tab:ksweep}
\small
\setlength{\tabcolsep}{6pt}
\begin{tabular}{lcccccc}
\toprule
 & \multicolumn{3}{c}{Tencent Beijing} & \multicolumn{3}{c}{GeoLife} \\
\cmidrule(lr){2-4}\cmidrule(lr){5-7}
$K$ & SFP$\downarrow$ & $|\Delta\beta|\!\downarrow$ & $|\Delta\kappa|\!\downarrow$ & SFP$\downarrow$ & $|\Delta\beta|\!\downarrow$ & $|\Delta\kappa|\!\downarrow$ \\
\midrule
4 (default) & \textbf{.003} & \textbf{.021} & \textbf{24}      & .003          & \textbf{.026} & 131              \\
5           & .007          & .078          & 146              & \textbf{.002} & .033          & 237              \\
6           & .004          & .059          & 407              & .004          & .058          & \textbf{102}     \\
\bottomrule
\end{tabular}
\end{table}

On Tencent Beijing $K\!=\!4$ wins SFP, $|\Delta\beta|$, and $|\Delta\kappa|$ simultaneously. On GeoLife the three configurations are clustered: $K\!=\!4$ wins $|\Delta\beta|$, $K\!=\!5$ marginally improves SFP, and $K\!=\!6$ marginally improves $|\Delta\kappa|$. We retain $K\!=\!4$ for its strict dominance on the larger-pool Tencent Beijing corpus.

\section{Architectural Components}
\label{appendix:arch}

This appendix specifies two architectural components mentioned in \S\ref{sec:method} as implementation details rather than ablated contributions.

\paragraph{Jacobian correction.}
The term $J_k(c) = \log \hat{\rho}_{k}\!\bigl(\log d(c_{i-1}, c)\bigr)$
in Eq.~\eqref{eq:cell} is an empirical correction that accounts for the
non-uniform density of candidate grid cells in log-distance space, including
effects from the distance transformation and grid geometry. Here,
$\hat{\rho}_{k}$ is a Gaussian kernel density estimate (KDE) of the empirical log-distance distribution between grid cells, with bandwidth derived from
$\sigma_k$ and a small lower-bound safeguard. It is precomputed once at model construction and serves as a
numerical safeguard against cell-density bias; we did not isolate its
contribution in the experiments.

\paragraph{Persona-conditioned generation via user latents.}
For generation we attach a learnable latent code
$z_u \in \mathbb{R}^{16}$ to each training user, mixed into the anchor
stream of every input token. Users are clustered into $P$ personas during
preprocessing, and a learnable Gaussian prior
$\mathcal{N}\!\bigl(\mu_p,\mathrm{diag}(\sigma_p^2)\bigr)$ is maintained
for each persona. A persona-conditioned regularization term encourages
each per-user code to remain close to its persona-specific prior. At
generation time, all experiments replay the trained user codes so that
synthetic and real populations share the same anchor distributions.

\end{document}